\documentclass[letterpaper, 10 pt, conference]{ieeeconf}  % Comment this line out if you need a4paper

\IEEEoverridecommandlockouts                              % This command is only needed if 
\usepackage{graphics} % for pdf, bitmapped graphics files
\usepackage{epsfig} % for postscript graphics files
\usepackage{mathptmx} % assumes new font selection scheme installed
\usepackage{times} % assumes new font selection scheme installed
\usepackage{amsmath} % assumes amsmath package installed
\usepackage{amssymb}  % assumes amsmath package installed
\usepackage{algorithm}
\usepackage{algorithmic}
\usepackage{subfigure}
\usepackage{multirow}
\let\labelindent\relax
\usepackage{enumitem}
\usepackage{pgfplots}
\usepackage{pgfplotstable}

\usepackage{scalerel}
\usepackage{tikz}
\usetikzlibrary{svg.path}
\definecolor{orcidlogocol}{HTML}{A6CE39}
\tikzset{
  orcidlogo/.pic={
    \fill[orcidlogocol] svg{M256,128c0,70.7-57.3,128-128,128C57.3,256,0,198.7,0,128C0,57.3,57.3,0,128,0C198.7,0,256,57.3,256,128z};
    \fill[white] svg{M86.3,186.2H70.9V79.1h15.4v48.4V186.2z}
                 svg{M108.9,79.1h41.6c39.6,0,57,28.3,57,53.6c0,27.5-21.5,53.6-56.8,53.6h-41.8V79.1z M124.3,172.4h24.5c34.9,0,42.9-26.5,42.9-39.7c0-21.5-13.7-39.7-43.7-39.7h-23.7V172.4z}
                 svg{M88.7,56.8c0,5.5-4.5,10.1-10.1,10.1c-5.6,0-10.1-4.6-10.1-10.1c0-5.6,4.5-10.1,10.1-10.1C84.2,46.7,88.7,51.3,88.7,56.8z};
  }
}
\newcommand\orcidicon[1]{\href{https://orcid.org/#1}{\mbox{\scalerel*{
\begin{tikzpicture}[yscale=-1,transform shape]
\pic{orcidlogo};
\end{tikzpicture}
}{|}}}}

\usepackage{hyperref} %<--- Load after everything else

\title{\LARGE \bf
ASVLite: a high-performance simulator for autonomous surface vehicles
}

\author{Toby Thomas \orcidicon{0000-0003-4664-5018}, %
        David M. Bossens \orcidicon{0000-0003-1924-5756} and %
        Danesh Tarapore \orcidicon{0000-0002-3226-6861}% <-this % stops a space
\thanks{
Authors are with the School of Electronics and Computer Science, %
University of Southampton, SO17 1BJ Southampton, U.K.}% 
}

\begin{document}

\maketitle
\thispagestyle{empty}
\pagestyle{empty}

\begin{abstract}

The energy of ocean waves is the key distinguishing factor of marine environments compared to other aquatic environments such as lakes and rivers. Waves significantly affect the dynamics of marine vehicles; hence it is imperative to consider the dynamics of vehicles in waves when developing efficient control strategies for autonomous surface vehicles (ASVs). However, most marine simulators available open-source either exclude dynamics of vehicles in waves or use methods with high computational overhead. This paper presents ASVLite, a computationally efficient ASV simulator that uses frequency domain analysis for wave force computation. ASVLite is suitable for applications requiring low computational overhead and high run-time performance. Our tests on a Raspberry Pi 2 and a mid-range desktop computer show that the simulator has a high run-time performance to efficiently simulate irregular waves with a component wave count of up to 260 and large-scale swarms of up to 500 ASVs. 

\end{abstract}
\section{Introduction}
Autonomous surface vehicles (ASVs), also known as unmanned surface vehicles (USVs), are vessels that operate on the surface of the water without any crew. A rapidly expanding market, driven by scientific, commercial and military interests, has resulted in ASVs been successfully deployed in a wide variety of missions ranging from environmental monitoring of oceanic waters to inspecting offshore maritime structures and performing security and patrol operations in coastal waters \cite{manley2008unmanned, papadopoulos20113d}. Importantly, swarms of these vessels have the potential to provide a high degree of spatio-temporal situational awareness of rapidly evolving marine and maritime disturbances occurring across large areas \cite{lonvcar2019heterogeneous}.

An essential tool required for the design and development of ASVs for oceanic environments is a vehicle dynamics simulator \cite{liu2016unmanned}. However, to the best of our knowledge, current open-source vehicle dynamics simulators for this domain are primarily designed for underwater vehicles. These do not simulate irregular ocean waves and the dynamics of vehicles in waves while waves may have a significant impact on the dynamic positioning and manoeuvrability of an ASV \cite{niu2018energy}. Moreover, wave forces also need to be accounted for when designing coordination strategies for swarms of ASVs, for instance, to synchronise long-range low-latency line-of-sight communication links between vehicles in a swarm.

Computing wave forces on an ASV has a significant computational overhead. Wave forces are computed by integrating the wave pressure along the wetted hull surface, which is computed as the intersection of geometries representing the hull surface and the instantaneous sea surface. This approach requires extensive computations due to approximating a surface integral and computing the geometric intersections of 3D objects. The computational overhead is even higher when simulating individual vehicles of a swarm due to the increased number of hull surfaces and the larger ocean surface encompassed by the entire swarm. 

One method to reduce computation is to simplify the hull geometry to an equivalent bounding box \cite{mendoncca2013kelpie} or to divide the hull into smaller segments and assume a constant waterline for each segment \cite{paravisi2019unmanned}. An alternative method to reduce computation and improve performance is to use a combination of (i) clustering of neighbouring facets over which wave forces are computed, (ii) parallelisation of wave force computation across these clusters, and (iii) a reduction in the number of instances the wave force computation is repeated in the simulation \cite{thakur2011real}.

A common thread in the above heuristics is the following: (i) the use of time-domain analysis for computing wave forces on the vehicle, i.e., integrating the wave pressure on the wetted hull surface at each time step of the simulation; and (ii) the use of a simplified hull mesh to reduce computational expense. However, simplifying the hull mesh alone does not provide a scalable solution; for example, when simulating individual vehicles of a swarm, the gains from reducing the complexity of each hull are negated by the higher number of hulls and the larger ocean surface to be simulated. While parallelisation may reduce the computation time, it does not reduce the computation itself, and high-degrees of parallelisation may not be supported onboard small-sized low-cost marine vehicles.

This paper describes ASVLite, a high-performance simulator which uses frequency-domain analysis to simulate the dynamics of an ASV in ocean waves while accounting for the effects of winds and currents. Instead of improving performance by simplifying the geometry of the vehicle, frequency-domain analysis takes advantage of the model of the irregular ocean waves to achieve high performance. Irregular ocean waves are modelled as a linear superposition of several regular component waves, and we assume that each regular component wave induces a periodic force on the ASV. The net force on the ASV is the superposition of the periodic forces induced by the component waves. The force induced by each component wave is computed by integrating the wave pressure along the hull and is reduced to a cosine function of the vehicle's position and time. Frequency-domain analysis delivers high performance because the computationally intensive operation of integrating the wave pressure along the wetted hull surface is performed only once per component wave, whereas in time-domain analysis the computation is repeated at each time step of the simulation.

ASVLite computes wave forces on the marine surface vehicle in two stages. The first stage computes wave forces on the vehicle for each component wave, and the second stage computes the net wave force on the vehicle by summing the instantaneous values of the cosine functions. Dividing the computation into two stages provides opportunities to reduce the computational overhead and improve run-time performance. The first stage of computation, which has a higher computational overhead, is performed before simulation of vehicle dynamics, thereby reducing the computation at run-time. Also, the performance of ASVLite becomes independent of the complexity of the vehicle's hull mesh since the number of cells in the geometry only affects the computations in the initial phase. Such splitting of computation into two phases also makes ASVLite ideal for simulation onboard an ASV since the first stage can be performed on a computer with higher computational capacity, and the results can be used for running the second stage on the ASV's onboard computer.

When simulating a swarm of ASVs, ASVLite takes advantage of multiple CPU cores, when available, by multithreading the computation for each ASV on parallel threads. ASVLite has been implemented with a clear and simple programming interface written in C programming language, making it easy to integrate with any existing or future software. The low run-time overhead of ASVLite makes it ideal for applications such as onboard simulations for behavioural adaptation through trial and error, and for applications that require high run-time performance such as in the simulation of a swarm of ASVs.
\section{Related work}

We consider the following features as essential in a marine vehicle simulator: (i) ability to simulate realistic ocean waves corresponding to a meteorologically given sea state; (ii) ability to simulate vehicle dynamics in waves in all six degrees of freedom;  and (iii) low computational overhead to provide high run-time performance and to enable applications onboard the vehicle. In this section, we review existing open-source marine vehicle simulators considering these features.

UWSim \cite{prats2012open} is a well-referred open-source hardware-in-the-loop simulator and provides a wide range of sensor modules and realistic rendering of the underwater environment. However, the hydrodynamic forces are computed outside the simulator in a MATLAB module, and therefore the simulator has a poor run-time performance. USVsim \cite{paravisi2019unmanned} is based on UWSim \cite{prats2012open} and provides additional features such as simulating forces due to wind, water current and waves. However, the wave force computation in USVsim is limited to hydrostatic forces, ignoring the hydrodynamic forces due to waves.

UUV Simulator \cite{manhaes2016uuv} and MARS \cite{tosik2014mars} were primarily developed for the simulation of underwater vehicles, but are also suitable for the simulation of a swarm of vehicles. Whereas UUV Simulator does not compute wave forces following the assumption that the vehicle operates outside the wave zone, MARS computes forces due to waves and water currents, but the wave force computation is limited to hydrostatic forces.  

Kelpie \cite{mendoncca2013kelpie} was developed for testing control algorithms for multi-robot systems of ASVs and aerial vehicles. Kelpie simulates ocean waves as regular waves since the simulator was developed with the assumption that irregular waves are not necessary for testing control algorithms. This assumption has been contradicted in Paravisi et al.  \cite{paravisi2019unmanned}, where accurate modelling of natural disturbances is considered essential, especially for small vehicles with low inertia, for developing efficient guidance, navigation, and control strategies. Also, the wave force computation in Kelpie is limited to hydrostatic forces. The source code of the simulator is not publicly available.  

Unlike the other simulators that we reviewed, the simulator developed by Thakur et al.  \cite{thakur2011real} can accurately simulate dynamics of an ASV in irregular ocean waves considering both the hydrostatic and hydrodynamic forces due to the waves. Although various heuristics were proposed to improve the run-time performance, the use of time-domain analysis for computing wave forces still makes the simulator computationally expensive and not suitable for achieving high run-time performance.

In summary, most marine vehicle simulators either ignore waves forces or limit the wave force computation to hydrostatic forces, ignoring the hydrodynamic forces due to waves. Although works such as that of Thakur et al. \cite{thakur2011real} model vehicle dynamics in waves, these models employ time-domain analysis for computing wave forces, a computationally expensive procedure repeated at each and every time step of the simulation, and thus are not suitable for achieving high run-time performance.  

\section{Methodology} \label{Methodology}

Here we propose a method for realistic simulation of ocean waves and their impact on ASVs. In ASVLite, the irregular ocean surface is modelled as a linear superposition of several regular waves with varying amplitude, frequency, and heading. This follows, to some extent, the earlier work in the realistic rendering of ocean waves \cite{frechot2006realistic,thon2002ocean}. To model the vehicle dynamics in waves, we assume that each regular component wave induces a periodic force on the ASV and the net wave force, $F_w$, on the vehicle at any instant of time is the linear superposition of forces due to each component wave.  

In ASVLite, the entire computation is divided into two phases. The first phase computes the properties of all component waves (amplitude, frequency, wave heading, phase) and the amplitude of the periodic force that each component wave exerts on the ASV's hull. The second phase computes the acceleration, velocity and position of the ASV by solving the equations of rigid body dynamics. Section \ref{Vehicle dynamics} gives an overview of the governing equation of rigid body dynamics of an ASV, and Section \ref{Computing wave force} details the computation of ocean waves and the  force that they exert on an ASV based on frequency-domain analysis.

\subsection{Dynamics of marine vehicle} \label{Vehicle dynamics}

The simulator computes the displacement, velocity and acceleration of the vehicle in 6 DoF for each time step based on the equations of rigid body dynamics:
\begin{equation}
  (M + M_A) a + C(v) v + K \Delta x = F_P + F_E\,, 
  \label{eq: rigid body dynamics}
\end{equation}
where $(M + M_A) a$ is the inertia force, $M$ is the mass matrix, $M_A$ is the added mass matrix, and $a$ is the acceleration vector. The added mass is computed using an empirical formula \cite{DNVGL}, assuming the hull shape as equivalent to an elliptical cylinder with a length of major axis equal to the length at waterline, length of minor axis equal to breadth at waterline and height of cylinder equal to the floating draught of the vehicle. $C(v) v$ is the hydrodynamic damping force, $C(v)$ is the damping matrix, and $v$ is the velocity vector. The hydrodynamic damping force acting on the vehicle is the sum of potential damping due to radiated wave, linear viscous damping due to skin friction and quadratic drag. At low speed (below 2~m/s), the potential damping and linear viscous damping is negligibly small and hydrodynamic damping can be considered equal to quadratic drag (\cite{fossen2011}, pp.126-130).  The drag force on the vehicle is computed based on an empirical formula \cite{DNVGL}, assuming hull geometry equivalent to an elliptical cylinder.  $K \Delta x$ is the hydrostatic restoring force, $K$ is the hydrostatic stiffness matrix, and $\Delta x$ is the displacement from the equilibrium floating attitude.  The excitation force acting on the vehicle $F_P + F_E$ is computed as the resultant of propeller force, $F_P$, and environmental force, $F_E$, which is the sum of the forces due to wave, wind and current. $F_w$ is the wave force acting on the vehicle, and its computation is described in detail in section \ref{Computing wave force}.

The instantaneous acceleration $a(t)$ of the vehicle at time step $t$ is computed from Eq. \ref{eq: rigid body dynamics}. The velocity $v(t)$ and position $x(t)$ of the vehicle are then computed by forward integration with a fixed time step size of $\Delta t$.

\subsection{Computing wave forces} \label{Computing wave force}

\subsubsection{Modelling ocean waves}

The irregular sea surface is considered a superposition of many regular waves and the state of the sea is defined using the Pierson-Moskowitz spectrum (\cite{stansberg2002specialist}, pp. 545-546), which is a single parameter spectrum based on wind speed as input and provides the correlation between wave frequency $f$ and variance $S(f)$, or wave energy (\cite{lewis1988principles},  p. 14). The Pierson-Moskowitz spectrum is defined as: 
\begin{equation}
  S(f) = \frac{A}{f^5} e^{\frac{-B}{f^4}}\,,
  \label{eq: Pierson-Moskowitz spectrum}
\end{equation}
where 
$A = \alpha g^2 (2 \pi)^{-4}$,  $B = \beta (2 \pi \frac{U}{g})^{-4}$,   $\alpha = 8.10 \times 10^{-3}$,  $\beta = 0.74$, and $U$ is the wind speed in $m/s$ measured at a height of 19.5 m above the  surface.

The Pierson-Moskowitz spectrum is a point spectrum, which represents a sea where all waves head in a single direction, and using it would simulate an irregular sea surface that is infinitely long crested. The sea is short crested because waves move in many different directions. Consequently, ASVLite converts the point spectrum to a directional spectrum  using the ITTC-recommended spreading function (\cite{hughes2010ship}, p. 4-29):
\begin{equation}
  G(\mu) = \begin{cases}
    \frac{2}{\pi} \cos^2(\mu), & \text{if }(\theta - \frac{\pi}{2}) \leq \mu \leq 
                                 (\theta + \frac{\pi}{2}) \\
    0, & \text{otherwise}
    \end{cases}\,,
  \label{eq: spreading function}
\end{equation}
where $\theta$ is the wind direction measured with respect to geographic  North. The equation for the resultant directional spectrum is:
\begin{equation}
  S(f, \mu) = S(f) G(\mu)\,.
  \label{eq: directional spectrum}
\end{equation}

For ASVLite, the continuous wave spectrum is converted to a discrete spectrum with frequency bands of uniform width and frequencies ranging from the minimum threshold frequency, $f_{0.1}$, to the maximum threshold frequency , $f_{99.9}$. The minimum and maximum threshold frequencies for Pierson-Moskowitz spectrum are computed as per ITTC recommendations as (\cite{stansberg2002specialist}, pp.545-546):
\begin{equation}
  f_{0.1} = 0.652 f_p\,, \text{ and } f_{99.9} = 5.946 f_p\,,
  \label{eq: minimum threshold frequency}
\end{equation}
where $f_p  =  (\frac{4B}{5})^{\frac{1}{4}}$ is the peak spectral frequency.

The discrete direction spectrum is used to generate a list of regular waves  such that each frequency band in the spectrum represents a regular wave, and  the area of the band in the spectrum is equal to the variance of the regular  wave. Amplitude, $\zeta_a$, of the regular wave can be computed from its variance,  $S(f)$, as (\cite{lewis1988principles}, p.12):
\begin{equation}
  \zeta_a = \sqrt{2 S(f)}\,.
  \label{eq: variance and amplitude}
\end{equation}
Wave elevation for a regular wave at position $(x,y)$ at time $t$ is computed as:  
\begin{equation}
  \zeta (x,y,t) = \zeta_a \cos[k(x \sin \mu + y \cos \mu) - \omega t +
  \epsilon]\,,
  \label{eq: regular wave}
\end{equation}
where $\zeta_a$ is the wave amplitude, $\omega$ is the circular frequency,  $k = \frac{\omega^2}{g}$ is the wave number, $\mu$ is the wave heading and  $\epsilon$ the phase angle. The sea surface elevation at any instant of time is  the sum of elevations of all regular component waves and is computed as:
\begin{equation}
  z(x,y,t) = \sum _{i} (\zeta_a)_i \cos[k_i(x \sin \mu_i + y \cos \mu_i) - 
                                        \omega_i t + \epsilon_i]\,. 
  \label{eq: surface elevation}
\end{equation}
The process to generate the component waves based on wave spectrum is described in Algorithm \ref{algorithm: component waves}.

\begin{algorithm}
\caption{Algorithm to generate the component waves.}
\begin{algorithmic}
\STATE Define $U$      // wind speed in m/s.
\STATE Define $\theta$ // wind direction in radians.
\STATE $g \leftarrow 9.81 $ // acceleration due to gravity in m/s$^2$.
\STATE $\alpha \leftarrow 8.10 \cdot 10^{-3}$ // as per Eq. \ref{eq: Pierson-Moskowitz spectrum}.
\STATE $\beta \leftarrow 0.74$ // as per Eq. \ref{eq: Pierson-Moskowitz spectrum}.
\STATE $A \leftarrow \alpha g^2 (2 \pi)^{-4}$ // as per Eq. \ref{eq: Pierson-Moskowitz spectrum}.
\STATE $B \leftarrow \beta (2 \pi \frac{U}{g})^{-4}$ // as per Eq. \ref{eq: Pierson-Moskowitz spectrum}.
\STATE $f_p \leftarrow (\frac{4B}{5})^{\frac{1}{4}}$ // as per Eq. \ref{eq: minimum threshold frequency}.
\STATE $f_{0.1} \leftarrow 0.652 f_p$ // as per Eq. \ref{eq: minimum threshold frequency}.
\STATE $f_{99.9} \leftarrow 5.946 f_p$ // as per Eq. \ref{eq: minimum threshold frequency}.
\STATE $waves \leftarrow$ new 2D array // array for component waves.
\STATE Define $n_{f}$ // number for frequency bands.
\STATE Define $n_{\mu}$ // number of heading directions.
\STATE $\Delta_f \leftarrow \frac{f_{99.9} - f_{0.1}}{n_f}$
\STATE $\Delta_{\mu} \leftarrow \frac{\pi}{n_\mu}$
\FOR{$\mu$ in range $(\theta - \frac{\pi}{2})$ to $(\theta + \frac{\pi}{2})$}
\FOR{$f$ in range $f_{0.1}$ to $f_{99.9}$}
\STATE $S \leftarrow (\frac{A}{f^5} e^{\frac{-B}{f^4}}) (\frac{2}{\pi} \cos^2 (\mu)) \Delta_f \Delta_{\mu}$ // as per Eq. \ref{eq: directional spectrum}.
\STATE $\zeta_a \leftarrow \sqrt{2 S}$ // as per Eq. \ref{eq: variance and amplitude}.
\STATE $\epsilon \leftarrow$ random number in range $[0,360]$ generated from a uniform distribution. 
\STATE // Generate and insert a regular wave in $waves$.
\STATE waves.insert\_wave($amplitude = \zeta_a$, $frequency = f$, $heading = \mu$, $phase = \epsilon$)
\ENDFOR
\ENDFOR
\end{algorithmic}
\label{algorithm: component waves}
\end{algorithm}

\subsubsection{Computation of wave force due to an irregular sea}
The wave force on the vehicle is the sum of the Froude-Krylov force and the diffraction  excitation force. The Froude-Krylov force is due to the pressure variation around the hull due to the wave and the diffraction excitation force is due to the modification of the incident wave due to presence of the vehicle. The diffraction excitation force is negligibly small due to the relatively small size of an ASV, which is on the order of 10 m or less, compared to wavelength, which is on the order of 100 m; therefore, the wave force is approximated equal to Froude-Krylov force (\cite{lewis1988principles}, p. 43). The Froude-Krylov force on the vehicle due to the irregular ocean surface, $F_{w}$, is computed as the sum of the Froude-Krylov force due to each component wave $F_{w_i}$: 
\begin{equation}
  F_{w} = \sum_{i} F_{w_i}\,.
  \label{eq: net wave force irregular sea}
\end{equation}
$F_{w_i}$ is computed by integrating the product of wave pressure with the wetted hull surface area, $dS$, as:
\begin{equation}
  F_{w_i} = \oint_{S} p_{i}(z) dS\,.
\label{eq: wave force}
\end{equation}
$p_{i}(z)$ is the wave pressure at depth $z$, defined as:
\begin{equation}
  p_{i}(z) = \rho g e^{k z} \cos[k(x \sin \mu + y \cos \mu) - \omega_{e_i} t]\,,
  \label{eq: wave pressure}
\end{equation}
where $\omega_{e}$ is the encountered wave frequency and is computed as:
\begin{equation}
  \omega_e = \omega - \frac{\omega^2}{g} U \cos(\mu - \phi)\,.
  \label{eq: encountered frequency}
\end{equation}
Since the vehicle dimensions are much smaller than the wavelength, it is reasonable to assume the wave pressure within the limits of the vehicle to vary linearly. Consequently, Eq. \ref{eq: wave force} for the wave forces acting on a marine vehicle in 6 DoF can be simplified as:
\begin{equation}
F_{w_i} = \begin{bmatrix}
F_{w_{surge,i}} \\
F_{w_{sway,i}} \\
F_{w_{heave,i}} \\
F_{w_{roll,i}} \\
F_{w_{pitch,i}} \\
F_{w_{yaw,i}} 
\end{bmatrix} = \begin{bmatrix}
(p_{fore,i} - p_{aft,i}) A_{x} \\
(p_{sb,i} - p_{ps,i}) A_{y} \\
p_{cog,i} A_z \\
(p_{sb,i} - p_{ps,i}) \frac{A_z}{2} \frac{B}{4} \\
(p_{fore,i} - p_{aft,i}) \frac{A_z}{2} \frac{B}{4} \\
(p_{fore,i} - p_{aft,i}) \frac{A_y}{2} \frac{L}{4}
\end{bmatrix}
\end{equation}

where $p_{cog}$ is the wave pressure at the centre of gravity of the ASV, $p_{aft}$ and $p_{fore}$ are wave pressure at a distance of $\frac{L}{4}$ from the centre of gravity of the ASV measured towards the aft and fore respectively, and $p_{sb}$ and $p_{ps}$ are wave pressure at a distance of $\frac{B}{4}$ from the centre of gravity of the ASV measured towards the starboard side and port side. $A_z$ is the waterplane area of the vehicle, $A_{x}$ is the transverse sectional area of the vehicle below the waterline at a distance of $\frac{L}{4}$ from the centre of gravity, and $A_{y}$ is the longitudinal profile area of the ASV below the waterline at a distance of $\frac{B}{4}$ from the centre of gravity. 

\section{Results}

\subsection{Validation of vehicle dynamics simulated with ASVLite}

\begin{figure}
\centering
  \begin{minipage}{0.24\textwidth}
    \centering 
    \subfigure[Side view.]{\includegraphics[width=1.0\linewidth]{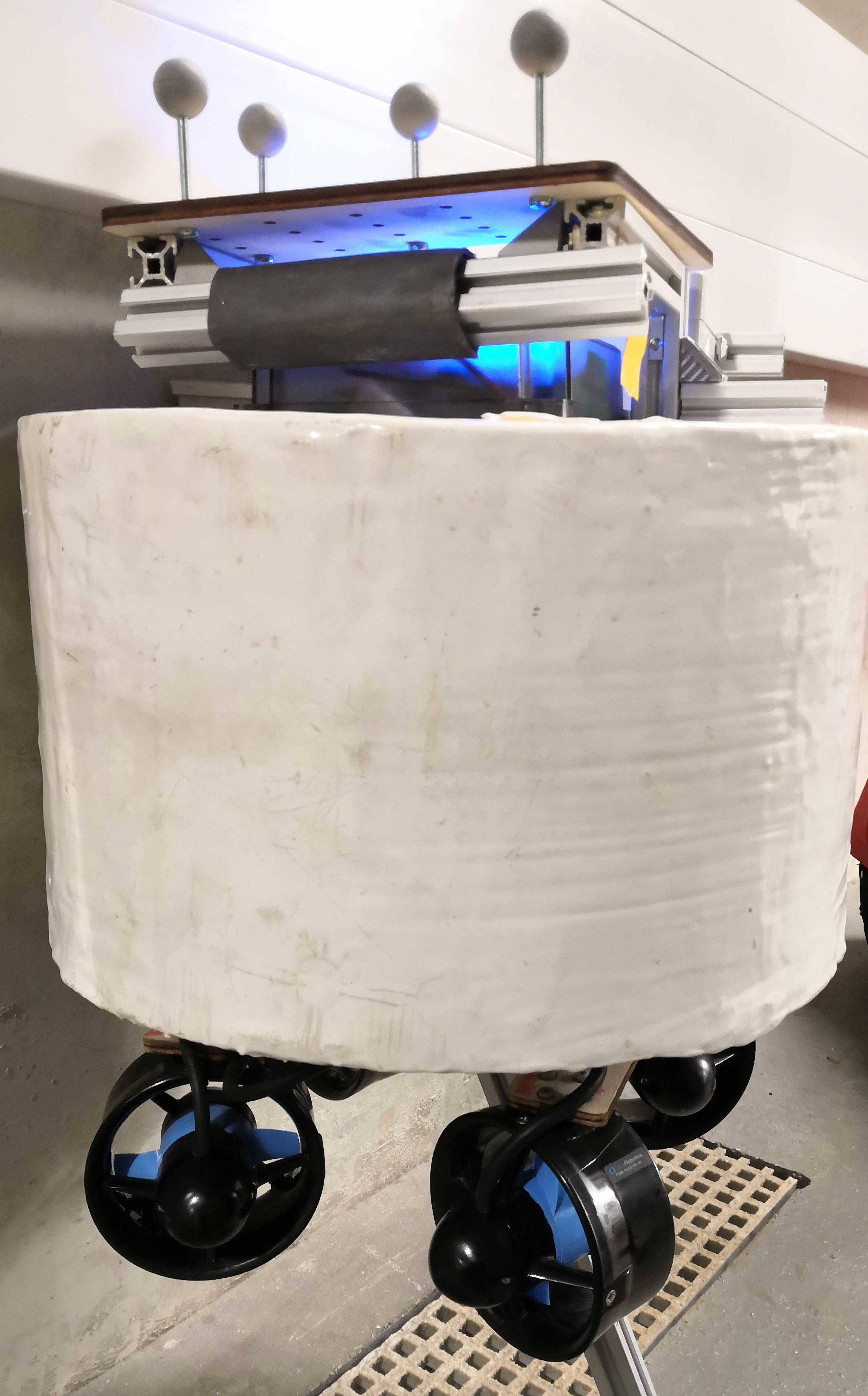}}
  \end{minipage}%
  \begin{minipage}{0.24\textwidth}
    \centering 
    \subfigure[Top view.]{\includegraphics[width=0.8\linewidth]{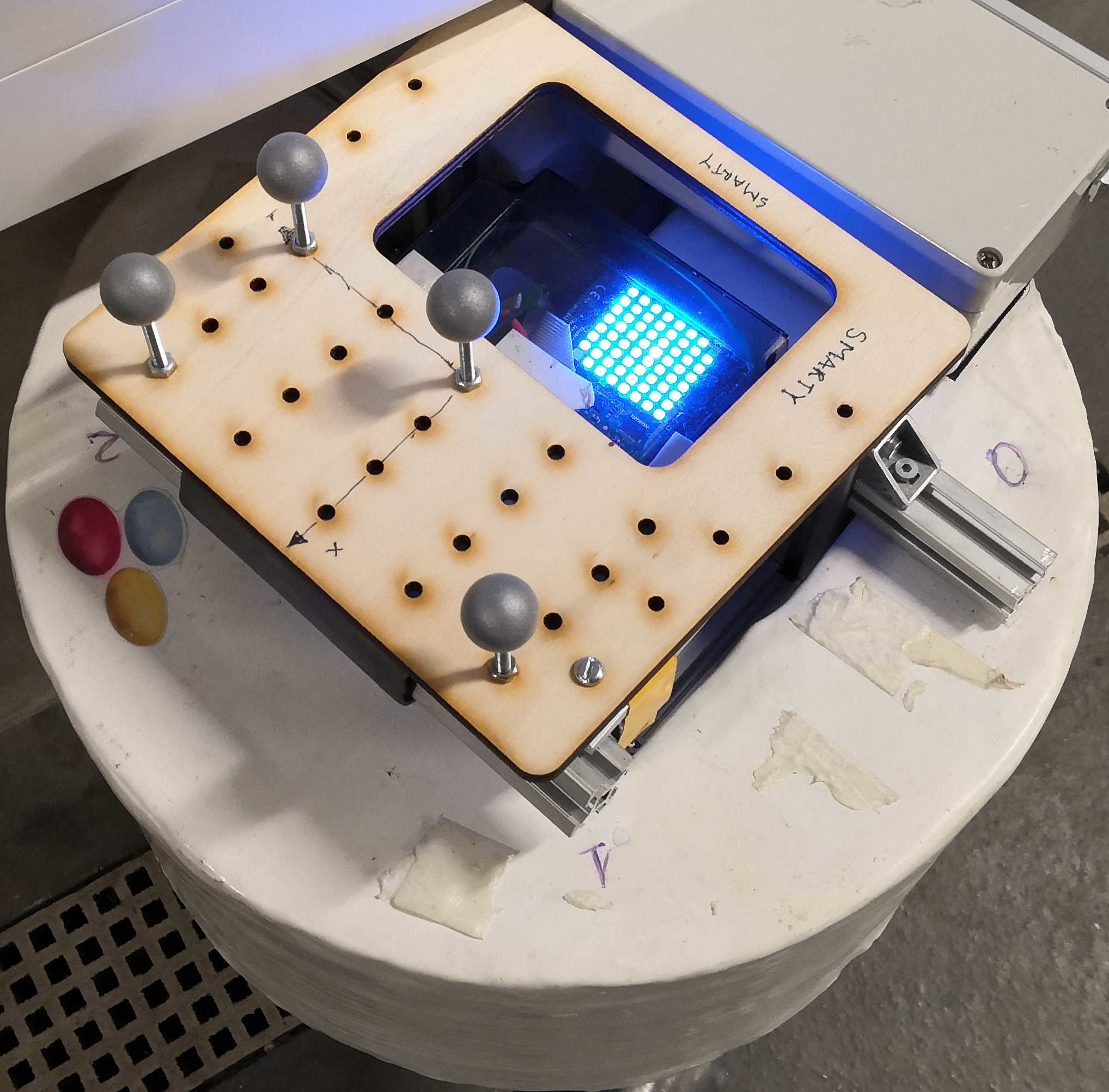}}\\ 
    \subfigure[Thruster configuration.]
      {\includegraphics[width=0.8\linewidth]{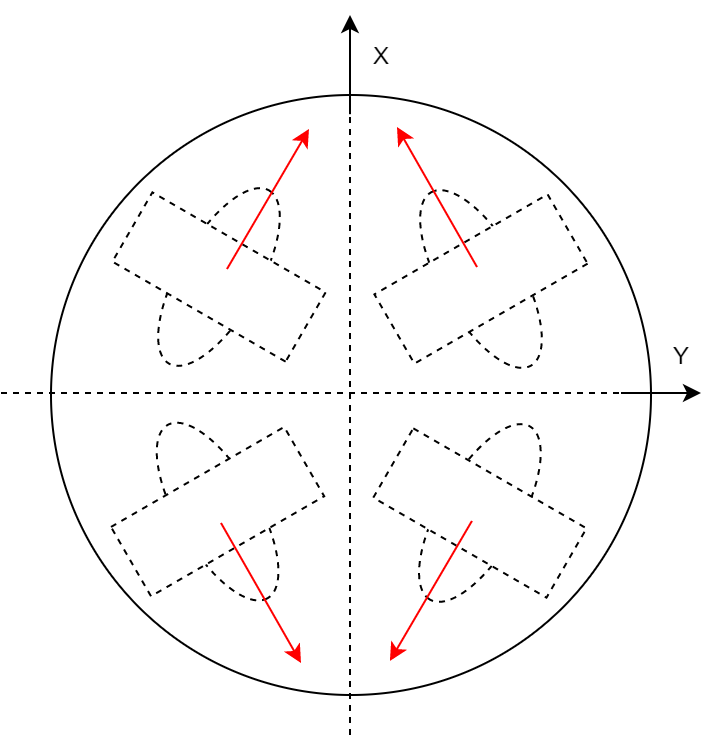}}
  \end{minipage}
\caption{Illustration of the SMARTY platform simulated.}
\label{fig: smarty}
\end{figure}

ASVLite was validated by comparing simulation results with data generated from running a remote-operated vehicle, SMARTY, in a towing tank capable of generating waves. SMARTY has a cylindrical shape and is equipped with four thrusters, as shown in Fig. \ref{fig: smarty}, with physical specifications shown in Table \ref{tab: smarty particulars}. A summary of the validation is presented for a wave of amplitude 6 cm in Table \ref{tab: towing tank vs ASVLite}. Not all the experimental details, such as the mass distribution and initial conditions, could be mimicked; however, we can confirm that the range of motion is of the same order.

\begin{table}[htbp]
\centering
\caption{Physical parameters of the SMARTY platform.}
\begin{tabular}{c c | c c }
\hline
Diameter  & 0.32 m & Height  &  0.21 m \\
Mass  & 8.4 kg & Draught  & 0.11 m \\
Thrusters  &  \multicolumn{3}{c}{BlueRobotics T100 Thruster} \\
\hline
\end{tabular}
\label{tab: smarty particulars}
\end{table}

\begin{table}[hbtp]
\centering
\caption{Comparison of heave, roll and pitch motion of SMARTY in a towing tank with the simulated motion in ASVLite for a wave of amplitude 6 cm. }
\begin{tabular}{c | c  c | c  c}
\hline
SMARTY      & \multicolumn{2}{c|}{Towing tank}  & \multicolumn{2}{c}{ASVLite} \\
motion            & Minimum        & Maximum                  & Minimum & Maximum \\
\hline
Heave (cm)  & -6.8       & 6.8                  & -7.8    & 7.6   \\
Roll (deg)  & -17.2      & 8.9                  & -13.5  & 14.8    \\
Pitch (deg) & -11.5      & 9.7                  & -4.1  & 12.3    \\
\hline
\end{tabular}
\label{tab: towing tank vs ASVLite}
\end{table}

To investigate vehicle motion trends, we simulated SMARTY to move forward for $50$~m in an open sea, while varying each of significant wave heights and wave headings in separate and independent experiments. The significant wave heights (in m), and wave headings were of $\{0.5, 0.625, 0.75 \dotsc 2.0 \}$ and $\{0^{\circ}, 11.25^{\circ}, 22.5^{\circ} \dotsc 360^{\circ} \}$, respectively. Each experiment was replicated $100$ times with different random seed values. Therefore, in total, $42900$ ($13$ significant wave heights $\times$ $33$ wave headings $\times$ $100$ replicates) experiments were performed. In each experiment, forward motion was generated by applying a constant force of $4$~N, while constraining yaw motion to achieve a constant wave heading. The vehicle was simulated with a fixed time step size of $40$~ ms. 

\begin{figure*}
\centering
  \subfigure[Heave motion trends.]
    {\includegraphics[width=0.32\textwidth]{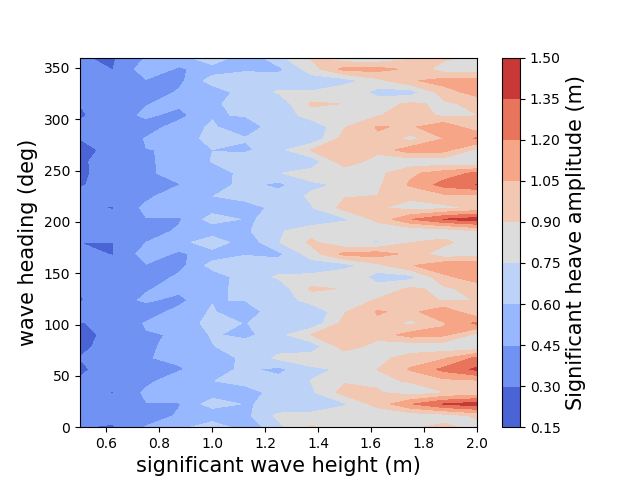}}
  \subfigure[Pitch motion trends.]
    {\includegraphics[width=0.32\textwidth]{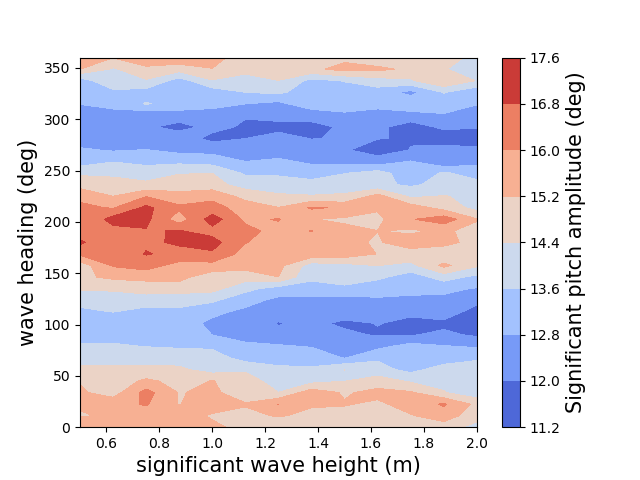}}
  \subfigure[Roll motion trends.]
    {\includegraphics[width=0.32\textwidth]{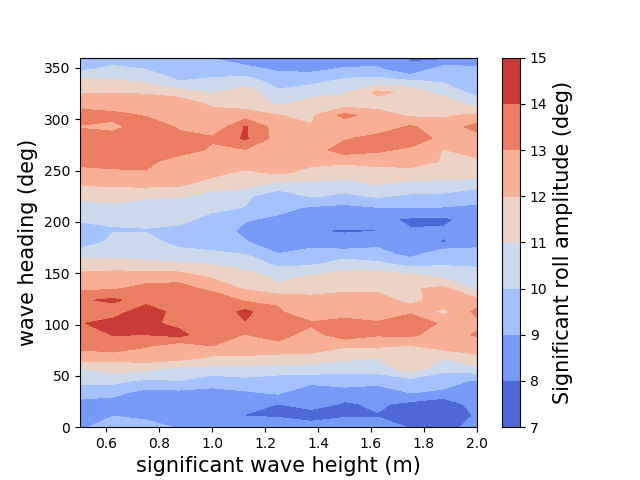}}
\caption{Simulated motion trends of SMARTY in an open sea for 13 different significant wave heights and 33 different wave heading angles, averaged over 100 replicates. A wave heading of $180^{\circ}$ corresponds to a head sea condition.}
\label{fig: motion trends}
\end{figure*}

Fig. \ref{fig: motion trends} shows the vehicle motion trends observed by plotting the mean of significant motion amplitudes for each combination of significant wave height and wave heading. As expected in a sea-going vessel, the heave motion of the simulated vehicle increased with an increase in wave height. However, the roll and pitch amplitude at first increased and then decreased with an increase in wave height; this is because as the sea state increases from calm to high, the spectral peak of the ocean waves moves towards a lower frequency. The waves become higher but also longer, making the waves less steep. Unlike heave motions of sea-going vessels, the simulated heave motions do not show any correlation with the wave heading, but this is due to the circular waterline of the simulated SMARTY platform. By contrast, the roll and pitch motions show a strong correlation with the wave heading. The pitch motion is highest in the head sea condition (wave heading at $180^{\circ}$), followed by the following sea condition (wave heading at $0^{\circ}$ or $360^{\circ}$) and lowest in the beam sea condition (wave heading at $90^{\circ}$ or $270^{\circ}$). The roll motion is highest in the beam sea condition and lowest in the head sea and following sea condition. In summary, the above noted trends of the dynamics of the simulated vessel exhibit characteristics similar to that observed of sea-going vessels in waves, thus demonstrating the high fidelity of ASVLite.

\subsection{Analysis of performance of ASVLite}

Experiments were also performed to estimate the performance of the ASVLite simulator. \textit{Performance is measured as the time required for completing the simulation and is expressed as the ratio of real-time to simulation-time}. Real-time is the time taken for the vehicle dynamics in the real world, and simulation-time is the time taken to complete the same dynamics in simulation. The experiments were run on a desktop computer, having a quad-core Intel Core i7-6700 CPU and 16GB of DDR4 2133MHz RAM, and on a Raspberry Pi 2, having a 900MHz quad-core ARM Cortex-A7 CPU and 1GB RAM.

The two key variables that influence the performance of ASVLite are the number of wave components used to generate the irregular sea surface and the size of the simulated swarm. Therefore, the forward motion experiments on the SMARTY platform were repeated while independently varying these two variables to assess their impact on performance. For all experiments, the simulation time step size was fixed at 40 ms, and the number of replicates was 100.

In the first set of experiments, with a single vehicle, the number of wave components was varied to observe its effect on performance (see Table \ref{tab: wave count - performance}). It is expected that a higher number of component waves while costly to simulate, defines a more realistic sea-surface with better short-crested waves. Our results indicate that ASVLite is capable of providing a high performance even with a larger number of component waves than recommended (75 wave components with 15 frequency bands per wave direction, see \cite{lewis1988principles}, p.13). When simulating with 75 component waves, the simulator performed 618x and 26x faster than real-time on the desktop computer and Raspberry Pi 2 platform respectively. 

In the second set of experiments, the performance of the simulator was tested by simulating marine vehicle swarms of varying size, while fixing the number of wave components at $75$ (see Table \ref{tab: swarm - performance}). 
%The desktop computer is expected to simulate the entire swarm whereas the Raspberry Pi 2, which may be used for onboard simulation, is expected to simulate only a subset of the swarm consisting of the immediate neighbours of the focal ASV. 
Simulations were performed using the following: (A) single-threading of the entire simulation process; (B) multi-threading with time synchronisation -- useful when simulating a tightly-coordinated swarm of ASVs or for visualising the resulting behaviour of the simulated swarm; and (C) multi-threading without time synchronisation -- useful when the ASVs in the swarm act largely independently of each other.
%for cases which require maximisation of the number of simulation per time.
Results from running ASVLite on the desktop indicate that the simulator can perform 280x faster than real-time when simulating a swarm of 10 vehicles, with the performance reducing to 7x for a large swarm of 500 vehicles. On the Raspberry Pi 2, the simulator performed 10x faster than real-time for a swarm of 10 vehicles and performed approximately at real-time speed when simulating a swarm of 100 vehicles.  

\begin{table}[hbtp]
\centering
\caption{Run-time performance of ASVLite averaged over 100 replicates, on a desktop computer and on a Raspberry Pi 2, for different number of regular wave components of the irregular sea surface.}
\begin{tabular}{c | c | c }
\hline
Component wave count                       & \multicolumn{2}{c}{Performance}                                       \\ 
(wave directions $\times$                  & \multicolumn{2}{c}{(real-time / simulation-time)}                     \\\cline{2-3}
 frequency bands)                          & Desktop  & Raspberry Pi 2    \\
\hline
15 ($3 \times 5$)                          & 3266x      & 132x                      \\
30 ($3 \times 10$)                         & 1648x      & 68x                        \\
75 ($5 \times 15$)                         & 618x       & 26x                        \\
135 ($9 \times 15$)                        & 342x       & 15x                        \\
195 ($13 \times 15$)                       & 237x       & 10x                        \\
260 ($13 \times 20$)                       & 178x       & 8x                          \\
\hline
\end{tabular}
\label{tab: wave count - performance}
\end{table}

\begin{table}[hbtp]
\centering
\caption{Run-time performance of ASVLite averaged over 100 replicates, on a desktop computer and a Raspberry Pi 2, for simulation of a swarm of ASVs using (A) single-threading, (B) multi-threading with time synchronisation, and (C) multi-threading without time synchronisation. Simulation of swarms of over 150 ASVs were not performed on the Raspberry Pi 2 as the simulation-time far exceeded real-time. %because  (1) the simulation-time will be greater than real-time, and (2) onboard simulation of swarms require only a subset of the swarm, comprising of the immediate neighbours of an ASV.  
}
\begin{tabular}{c | c c c | c c c}
\hline
                    & \multicolumn{6}{c}{Performance}                                       \\ 
Number of ASVs      & \multicolumn{6}{c}{(real-time / simulation-time)}                     \\\cline{2-7}
in simulated swarm  & \multicolumn{3}{c|}{Desktop}  & \multicolumn{3}{c}{Raspberry Pi 2}    \\
                    & A     & B     & C             & A     & B     & C                     \\                      
\hline
10                  & 67x   & 150x  & 280x          & 3x    & 7x    & 10x                   \\
50                  & 14x   & 44x   & 64x           & 1x    & 2x    & 2x                    \\
100                 & 7x    & 23x   & 32x           & 0.3x  & 1x    & 1x                    \\
150                 & 5x    & 16x   & 22x           & 0.2x  & 0.6x  & 0.7x                  \\
200                 & 3x    & 12x   & 16x           & -     & -     & -                     \\
250                 & 3x    & 9x    & 13x           & -     & -     & -                     \\
500                 & 1x    & 5x    & 7x            & -     & -     & -                     \\
\hline
\end{tabular}
\label{tab: swarm - performance}
\end{table}

\section{Conclusion and future works}

ASVLite achieves high run-time performance by using frequency-domain analysis for simulating the dynamics of the vehicle in waves. Its two-stage computation, with the wave force computation in the first stage, reduces the computational overhead during the simulation of vehicle dynamics, which is in the second stage, thereby resulting in high run-time performance independent of the complexity of the hull mesh geometry. Furthermore, ASVLite takes advantage of multi-threading when simulating a large swarm of ASVs where the simulation of each ASV runs in a parallel thread.

ASVLite is implemented based on the assumption that the irregular sea surface is composed of many regular waves and the net wave force at any instant of time is a linear summation of wave force due to each component wave. The wave pressure variation along the hull is approximated as linear due to the relatively small size of an ASV, on the order of 10~m or less, compared to the sea wave lengths, on the order of 100~m or more. These assumptions hold for most use cases of current ASVs which are slow-moving displacement vessels with speeds less than 10 knots and operating predominantly in mild sea conditions. High-speed simulation with high fidelity of non-linear systems such as the dynamics of a high-speed planning craft or sea states with wave breaking is currently not possible with ASVLite but may be achieved in future work by combining it with computational fluid dynamics (CFD) based approaches.

\section{Supplementary information}

The simulator is publicly available as open-source at https://github.com/resilient-swarms/ASVLite.git.
\section{Acknowledgement}
The work was funded in part by an EPSRC New Investigator Award (EP/R030073/1) to DT. The remote-operated vehicle, SMARTY, and the towing tank facility are part of the Department of Civil, Maritime and Environmental Engineering at University of Southampton, United Kingdom, and the authors would like to thank Dr Blair Thornton and his team of researchers for providing SMARTY data for validating the simulator.

\bibliographystyle{IEEEtran}
\bibliography{reference}

\end{document}